%% file: main.tex
\documentclass{article}

\usepackage{microtype}
\usepackage{graphicx}
\usepackage{booktabs}
\usepackage{multirow}
\usepackage{makecell}
\usepackage{hyperref}

\usepackage[accepted]{icml2026}
\usepackage{amsmath, amssymb, amsfonts}
\usepackage[dvipsnames]{xcolor}
\usepackage{placeins}
\usepackage{float}
\usepackage{enumitem}
\input{math_commands}

\icmltitlerunning{Emergent Symbolic Structure in Health Foundation Models}

\begin{document}
\raggedbottom

\twocolumn[
\icmltitle{Emergent Symbolic Structure in Health Foundation Models:\\
Extraction, Alignment, and Cross-Modal Transfer}

\icmlsetsymbol{equal}{*}

% Author block hidden for blind review
\begin{icmlauthorlist}
\icmlauthor{Gajendra Katuwal}{apple}
\icmlauthor{Advait Koparkar}{apple}
\icmlauthor{Salar Abbaspourazad}{apple}
\icmlauthor{Anshuman Mishra}{apple}
\icmlauthor{Sarvesh Kirthivasan}{apple}
\end{icmlauthorlist}

\icmlaffiliation{apple}{Apple}

\icmlcorrespondingauthor{Gajendra Katuwal}{gkatuwal@apple.com}
% \icmlcorrespondingauthor{Advait Koparkar}{a\_koparkar@apple.com}
% \icmlcorrespondingauthor{Salar Abbaspourazad}{salarabb@apple.com}
% \icmlcorrespondingauthor{Anshuman Mishra}{a\_mishra@apple.com}
% \icmlcorrespondingauthor{Sarvesh Kirthivasan}{skirthivasan@apple.com}

\icmlkeywords{mechanistic interpretability, health foundation
models, symbol extraction, cross-modal transfer}

\vskip 0.3in
]

\printAffiliationsAndNotice{\textit{Authors ordered by contribution.}}

\begin{abstract}
We show that information can be transferred post-hoc across independently trained health foundation models (FMs), each pretrained on ${\sim}$20M minutes of wearable sensor data from ${\sim}$172K participants,
 by aligning their data-dependent coordinate systems.
From frozen embeddings we extract candidate symbol-like components using linear decomposition methods, and align them across models with simple linear maps.
Aligned symbols associate selectively with health conditions and physiological attributes, with associations similar across modalities and architectures. 
A classifier trained on one model's symbols and applied to another retains more than 95\% of its in-domain performance, with similar retention in both directions.
Overall, our results indicate that independently trained health FMs converge toward a common representation of the same underlying physiology.
\end{abstract}

% ----------------------------------------------------
\section{Introduction}
Health FMs trained on wearable
biosignals~\cite{spathis2020selfsupervised, abbaspourazad2023wearable,
abbaspourazad2024accel, narayanswamy2024scaling,
pillai2024papagei, gu2025cardiac} have demonstrated that
health-relevant structure can be learned from large amounts of
unlabeled sensor data. However, like other deep learning systems,
they lack explicit symbolic structure for transparent,
structured reasoning~\cite{kahneman2011thinking}. Bridging this
gap is a central goal of neurosymbolic
AI~\cite{garcez2020neurosymbolic}, and is especially pressing in
healthcare, where interpretability and structured decision-making
are critical. Even partial progress, surfacing symbol-like
components from learned representations, would aid
interpretability and unlock new capabilities such as cross-modal
transfer. A natural question is whether such symbols must be
imposed externally, or whether they already emerge within the
learned representations and can be surfaced post-hoc.

Recent work provides theoretical motivation to expect that
interpretable, shared structure exists within neural embeddings.
The Linear Representation Hypothesis
(LRH)~\cite{elhage2024linear, park2024linear}, supported by
mechanistic interpretability
work~\cite{olah2020zoom, elhage2022mathematical}, posits that
neural networks encode individual concepts as linear directions
in representation space.
The Universality Hypothesis~\cite{li2016convergent, olah2020zoom,
chughtai2023toy} and the Platonic Representation
Hypothesis~\cite{huh2024platonic} go further, suggesting that
networks trained under differing architectures, modalities, and
objectives converge toward similar latent structures, implying
that symbol-like components may be partially shared across
modalities despite very different sensing mechanisms.

Motivated by these hypotheses, we ask:
 \begin{quote}
\textit{Do health FM embeddings admit a symbol-like
decomposition that is consistent across modalities? If so, do the
resulting components carry interpretable health meaning, and can
they support cross-modal transfer without joint training?}
\end{quote}

Such shared symbolic structure is not directly visible in raw
embeddings, where latent factors are entangled
across coordinates~\cite{hinton1986distributed,
bengio2013representation} and feature superposition compounds
the problem~\cite{olah2020zoom, elhage2022superposition}.
We address this by re-expressing \emph{frozen embeddings} in data-dependent coordinate systems 
whose components are more interpretable and comparable across models (\S\ref{sec:method}).
 We use \emph{symbol} as an operational term for a component with
selective health associations that can be extracted, aligned, and
transferred across modalities. Like classical
symbols~\cite{newell1976computer}, these are manipulable units
of knowledge, but grounded in continuous embedding geometry
rather than logical predicates. We evaluate them empirically
as candidate symbol-like factors, not as symbols in the strong
cognitive sense.

% \subsection{Contributions}
\paragraph{Contributions}

We introduce a unified framework for post-hoc symbol extraction,
alignment, and transfer from frozen embeddings without joint
training (\S\ref{sec:method}). To our knowledge, this is the
first work to study emergent symbolic structure in health
FMs and at this scale (${\sim}$20M minutes,
${\sim}$172K participants). Using this framework we show:

\begin{enumerate}
\item Extracted components exhibit selective associations with
health conditions and physiological targets, and these associations
are partially shared across sensing modalities and architectures
(\S\ref{sec:symbol_alignment}, \S\ref{sec:symbol_meaning}).

\item Embeddings from different modalities share a subspace (as defined in \S\ref{sec:symbol_alignment_method}) rich
in physiological information that simple linear maps can recover,
enabling cross-modal transfer without joint training from only
a fraction of the paired cohort
(\S\ref{sec:symbolic_transfer}).
\end{enumerate}

% ----------------------------------------------------

\section{Related Work}

Our framework connects to two lines of research: decomposing neural
representations into interpretable units, and aligning
representations across domains.

\paragraph{Interpretability through decomposition.}
Sparse autoencoders recover monosemantic features from polysemantic
neurons~\cite{bricken2023monosemanticity, cunningham2024sparse},
scaling to frontier models~\cite{templeton2024scaling, gao2024scaling}, extending to causal circuits~\cite{marks2024sparse} and
multi-dimensional features~\cite{engels2024nonlinear}.
\cite{yamagiwa2023ica} showed that ICA applied to word and image
embeddings reveals shared geometric structure across modalities.
Concept-based methods explain models through predefined
directions~\cite{kim2018tcav, koh2020concept, oikarinen2023labelfree}
or population-level representation
engineering~\cite{zou2023representation, marks2024geometry}.
Our work differs in that we study health FM embeddings,
require no concept supervision, and discover symbols from frozen
embeddings using classical linear decompositions evaluated
post-hoc against clinical attributes.

\paragraph{Cross-modal alignment and transfer.}
Most cross-modal alignment methods require joint
training~\cite{radford2021clip}. Post-hoc alternatives include model stitching, which learns a linear layer
between intermediate representations~\cite{bansal2021stitching},
and relative representations, which re-express embeddings as
similarities to shared anchor
points~\cite{moschella2023relative}. Comparison measures such as
Centered Kernel Alignment (CKA)~\cite{kornblith2019similarity} quantify representational
similarity but do not produce a mapping for transfer; for a
broader taxonomy see~\cite{sucholutsky2023alignment}.
Our approach operates on frozen embeddings through explicit
symbol alignment (one-to-one matching or linear map), enabling
both transfer and interpretability within the same framework.

% ----------------------------------------------------
\section{Method}
\label{sec:method}

\begin{figure*}[t]
\begin{center}
\centerline{\includegraphics[width=\textwidth]{%
  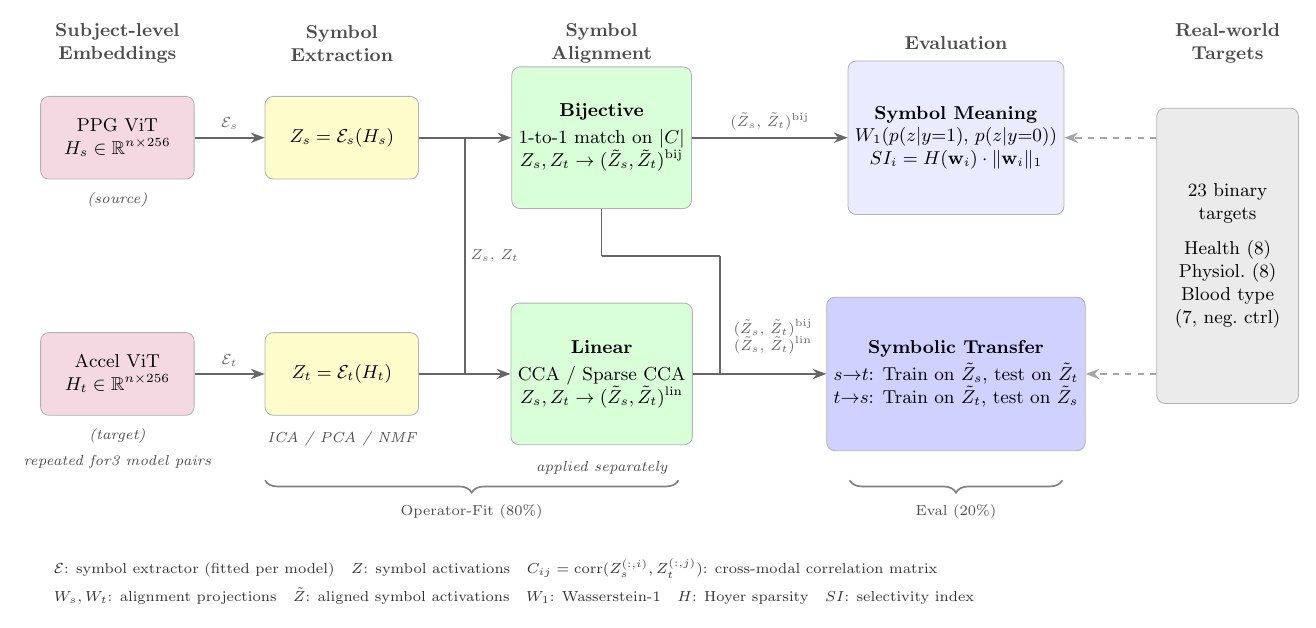}}
\caption{Overview of the symbolic framework. Frozen embeddings are
re-expressed as symbol activations via linear decompositions (ICA, PCA,
or NMF), then aligned across models using bijective matching
(Hungarian algorithm on $|C|$) or CCA. Aligned symbols are
evaluated for selective meaning (W1 distance against 23 targets)
and cross-modal transfer (classifier trained on source, tested on
aligned target). All operators are fitted on an 80\% operator-fit
split of a 30K held-out cohort; metrics use the remaining 20\%.}
\label{fig:method_overview}
\end{center}
\vskip -0.2in
\end{figure*}

\subsection{Dataset and Model Pretraining}
We use three separately trained FMs, each with its
own architecture, modality, and training objective rather than a
single multi-modal model. Each is trained on approximately 20 million
minutes of unlabeled sensor data collected under informed consent from the same
$\sim$172 thousand participants in the Apple Heart and Movement Study
(AHMS)~\cite{macrae2021ahms}:
\begin{itemize}[nosep]
    \item  Vision Transformer trained on PPG (PPG~ViT)
    \item  EfficientNet trained on PPG (PPG~EfficientNet)
    \item  Vision Transformer trained on accelerometer (Accel~ViT)
\end{itemize}
See~\cite{abbaspourazad2023wearable} and~\cite{abbaspourazad2024accel}
for architecture and pretraining details. Note that the Accel~ViT
used here is the uni-modal encoder trained solely on accelerometer
data, not the knowledge-distilled variant
from~\cite{abbaspourazad2024accel}.

Each model produces a $d{=}256$-dimensional embedding per 1 min input window.
We aggregate these into subject-level embeddings by averaging across
all windows for a given participant, yielding one vector
$h \in \mathbb{R}^{d}$ per subject per model.
This averaging discards within-subject temporal structure in favor of
a stable per-subject representation suitable for population-level
analysis.
Symbol extraction, alignment, meaning analysis, and transfer
experiments all use a held-out cohort of 30 thousand subjects not seen during
pretraining. We split this cohort 80/20 into an operator-fit set and an evaluation set.
Symbol operators and downstream classifiers are
fitted on the operator-fit set. All reported metrics use the evaluation
set with frozen operators.

\subsection{Symbolic Framework}
The framework consists of two steps: \emph{symbol extraction},
which re-expresses each model's embeddings as candidate symbols,
and \emph{symbol alignment}, which maps symbols across models.

\subsubsection{Symbol Extraction}
We view symbol extraction as applying an \emph{extraction operator}
$\mathcal{E}$ to frozen embeddings, re-expressing each model in a new
component representation whose factors are candidate symbols. Each
operator acts as a \emph{representation lens}: a different way of
viewing the same embedding space rather than a competing method, with
the choice of extractor depending on the downstream use case. In this
paper, we focus on linear decomposition methods for symbol extraction.

Let $H_m \in \mathbb{R}^{n \times d}$ denote the matrix of
subject-level embeddings from model $m$, where $n$ is the number
of subjects and $d$ is the embedding dimensionality ($d{=}256$).
A linear decomposition represents these embeddings as coefficients over
a learned symbol dictionary:
\begin{equation}
H_m \approx Z_m S_m^{\top},
\end{equation}
where $S_m \in \mathbb{R}^{d \times k}$ contains the candidate symbol
components and $Z_m \in \mathbb{R}^{n \times k}$ contains their
activations across subjects. In this work we use $k=d=256$. More
generally, a symbol extractor $\mathcal{E}_m$ maps embeddings to
activations,
\begin{equation}
Z_m = \mathcal{E}_m(H_m).
\end{equation}

We instantiate $\mathcal{E}_m$ using three symbol extractors:
Principal Component Analysis (PCA), orthogonal directions capturing
maximal variance; Independent Component Analysis (ICA), components
capturing statistically independent sources; and Non-negative Matrix
Factorization (NMF), yielding parts-based, additive components. These
methods impose distinct structural constraints on the components and
activations (Table~\ref{tab:symbol_extractors}), offering complementary
lenses on the symbolic structure of representation spaces.

In all cases, the extracted symbols are fixed linear components of the
embedding space, and $Z_m$ denotes the corresponding activation or
coefficient matrix. Thus these methods surface linearly recoverable
structure in the frozen embeddings. The framework also accommodates nonlinear extractors such as sparse
autoencoders; extending the analysis to these methods is left to future
work (Supplementary~\ref{sec:extraction_methods}).

\subsubsection{Symbol Alignment}
\label{sec:symbol_alignment_method}
Given symbol activations
$Z_{\text{source}}, Z_{\text{target}} \in \mathbb{R}^{n \times d}$ after symbol extraction
from two domains, alignment finds operators $W_s$ and $W_t$ that
map both into a shared coordinate system where corresponding
components are maximally correlated.
We use two alignment techniques.

\paragraph{(1) Bijective alignment.}
This technique finds a one-to-one matching that maximizes
pairwise similarity, preserving individual symbol identity
for interpretability and meaning analysis.
We compute a similarity matrix
\begin{equation}
C_{ij} = \mathrm{corr}\big(Z_{\text{source}}^{(:,i)},\, Z_{\text{target}}^{(:,j)}\big),
\end{equation}
where $Z^{(:,i)}$ denotes the activation of the $i$th symbol across
all subjects.
The optimal one-to-one assignment is found via the Hungarian
algorithm~\cite{kuhn1955hungarian} on $|C|$, yielding
$W_s$ and $W_t$ that establish bijective correspondence between symbols of the source and target domains, with polarity corrections ensuring all matched pairs are positively correlated. 
After Hungarian matching on $|C|$, we flip the sign of any target
component with $C_{ij}<0$ so each matched pair is positively correlated.
We keep every matched pair and use $|C_{ij}|$ as its correlation strength.

\paragraph{(2) Linear alignment.}
This technique relaxes the one-to-one constraint, allowing
shared structure to span multiple dimensions. It is better suited
for cross-modal transfer where maximizing overall alignment
matters more than preserving individual symbol identity. Canonical Correlation
Analysis (CCA)~\cite{hotelling1936cca} finds linear
projections $W_s, W_t$ of each domain's symbols into a shared
canonical space by maximizing cross-domain correlation:
\begin{equation}
 \begin{split}
 (W_s^{*},\, W_t^{*}) = \arg\max_{W_s,\, W_t} \;
 & \mathrm{tr}\big(W_s^\top \Sigma_{st}\, W_t\big) \\
 & - \lambda \big(\|W_s\|_1 + \|W_t\|_1\big),
\end{split}
\end{equation}
subject to $W_s^\top \Sigma_{ss}\, W_s = I$ and
$W_t^\top \Sigma_{tt}\, W_t = I$,
where $\Sigma_{st}$ is the cross-covariance between symbol
activations and $\Sigma_{ss}$, $\Sigma_{tt}$ are their respective
covariance matrices. Standard CCA sets $\lambda = 0$ and increasing $\lambda$
enforces sparser projections known as sparse CCA
~\cite{witten2009penalized}, constraining each
canonical direction to depend on only a few source symbols.
Concretely, CCA returns linear maps $W_s, W_t$ such that the aligned coordinates
 $Z_s W_s$ and $Z_t W_t$ are maximally correlated coordinate-by-coordinate. 
 When we refer to a \emph{shared subspace} in what follows, we mean the span of these correlated canonical directions. This is a property of the aligned coordinates,
  not a claim that the original embedding spaces share a canonical, model-independent subspace.

\subsection{Evaluation}
\label{sec:evaluation}

\input{tables/target_groups_table}

We evaluate the symbolic framework against a set of target
labels derived from participant metadata in the AHMS, organized
into three groups comprising 23 binary targets
(Table~\ref{tab:target_groups}).
Health condition labels are constructed from participants'
self-reported survey responses following the protocol
in~\cite{abbaspourazad2023wearable}; we aggregate related items
into eight condition categories (Supplementary
Table~\ref{tab:label_definitions}). Physiological markers (sex,
age, BMI) are from enrollment data. Blood type is user-reported
via the Apple Health app profile.
Health conditions and physiological markers serve as
\emph{positive controls} with expected physiological signal;
blood type serves as a \emph{negative control} with no expected
signal in either modality.

\subsubsection{Quantifying Symbol Meaning}
We quantify the association of a symbol to a target by measuring
the groupwise difference in activation along that symbol
direction using the Wasserstein-1 (W1)
distance~\cite{villani2009optimal}, with Cohen's $d$ as a
robustness check (see
Supplementary~\ref{sec:symbol_meaning_quantification} for
details). Computing W1 for each symbol-target pair yields a
condition profile $\mathbf{w}_i \in \mathbb{R}^{T}$ per
symbol $i$.

A high-quality symbol should associate strongly with a small
number of targets rather than responding diffusely to many.
We define the \emph{Selectivity Index} to capture both the
shape and magnitude of this profile:
\begin{equation}
\mathrm{SI}_i = H(\mathbf{w}_i) \cdot \|\mathbf{w}_i\|_1
\label{eq:selectivity_index}
\end{equation}
where $H$ is the Hoyer sparsity~\cite{hoyer2004nonnegative}
(shape) and $\|\cdot\|_1$ is the L1 norm (magnitude).
See Supplementary~\ref{sec:si_details} for derivation details.

\subsubsection{Symbolic transfer learning}
\label{sec:symbolic_transfer_method}
Having established that aligned symbols carry physiological
meaning, we quantify how well this structure transfers across
domains. We train a logistic regression classifier $f_\theta$
on source-domain symbols to predict target labels $y$, then
apply it to aligned target-domain symbols (subjects are paired,
so the same $y$ serves as ground truth):
\begin{align}
\theta^{*} &= \arg\min_{\theta}\;
\mathcal{L}\Big(f_\theta\big(
\underbrace{
\mathcal{E}_{\text{source}}(H_{\text{source}})\,W_s
}_{\tilde{Z}_{\text{source}}}
\big),\,
y\Big),
\label{eq:theta_star} \\
\hat{y}_{\text{target}} &=
f_{\theta^*}\!\Big(
\underbrace{
\mathcal{E}_{\text{target}}(H_{\text{target}})\,W_t
}_{\tilde{Z}_{\text{target}}}
\Big).
\label{eq:transfer}
\end{align}
For transfer, both the alignment operators and source-domain
classifiers are fit only on the operator-fit split; all AUCs are
computed on held-out evaluation subjects.
We use AUC (area under the receiver operating characteristic curve) as the primary transfer metric
(see Supplementary~\ref{sec:auc_justification} for justification).
We define \emph{transfer retention} as
\begin{equation}
\text{Transfer Retention} = \frac{\text{AUC}_{\text{transfer}}}
                       {\text{AUC}_{\text{in-domain}}},
\label{eq:retention}
\end{equation}
where $\text{AUC}_{\text{transfer}}$ is the AUC of the source
classifier applied to aligned target symbols and
$\text{AUC}_{\text{in-domain}}$ is the AUC when both training and
evaluation use the same domain. A value near 100\% indicates
nearly lossless transfer. We evaluate transfer separately for
positive-control targets (health conditions, physiological markers)
and the negative-control group (blood type), expecting the former to
transfer and the latter to remain near chance.

\subsubsection{Transfer efficiency}
\label{sec:transfer_efficiency_method}
A practical question for post-hoc alignment is how many paired
subjects are needed to learn a useful mapping between independently
trained models. If the shared structure is simple, few paired subjects may suffice.
Whereas if it is complex, more pairs will be needed.
We evaluate this by sweeping the training fraction
$\rho \in \{0.1, 0.2, \ldots, 1.0\}$ of participant overlap between
domain pairs and measuring transfer retention as a function of
$\rho$.

% ----------------------------------------------------
\section{Results}
\input{results_symbol_alignment}

\input{results_symbol_meaning_analysis}
\input{results_symbolic_transfer}

\section{Discussion}
\label{sec:discussion}

\paragraph{A shared, physiologically meaningful structure across modalities.}
Regardless of which extractor or modality pair we use, CCA
recovers near-identical cross-modal correlation
(Figure~\ref{fig:symbol_alignment}c).
This suggests that the coordinates CCA recovers, the shared
subspace as defined in \S\ref{sec:symbol_alignment_method}, are largely invariant to
modality pairing and extraction basis~\cite{lyu2022understanding}.
Beyond this geometric alignment, aligned symbols carry consistent
health associations across modalities, in line with the high
transfer retention ($>$98\%).
This shared structure persists even for raw embeddings, suggesting
that extraction sharpens rather than creates it, and that different
extractors simply offer complementary lenses on the same structure.
We read this as each modality offering a view of the same underlying
physiological system, consistent with the Platonic Representation
Hypothesis~\cite{huh2024platonic}: independently trained models can
converge toward compatible latent structure when trained on different
observations of the same process.

\paragraph{The cross-modal alignment is simple, with few effective degrees of freedom.}
Sparse CCA shows that much of
the alignment is preserved even when many projection weights are driven
to zero (Figure~\ref{fig:symbol_alignment}d), and transfer retention
saturates with only a fraction of the paired cohort
(\S\ref{sec:symbolic_transfer}). These results suggest that the
alignment recovered in our data has few effective degrees of freedom.
The similar retention in both transfer directions further supports this
interpretation. Both modality-specific maps, $W_s$ and $W_t$, recover
the downstream-predictive coordinates with comparable fidelity. Much of
the transferable signal therefore appears to lie in a shared predictive
component, so alignment mainly brings corresponding coordinates into
register rather than learning a full cross-modal translation. 

\paragraph{A parallel to cross-lingual word embeddings.}
Independently trained
word embedding spaces can often be brought into correspondence by
simple linear maps, often approximately orthogonal and structure
preserving~\cite{xing2015normalized,smith2017offline,artetxe2016learning}.
This was first shown to enable word translation from only a small seed
dictionary~\cite{mikolov2013exploiting}, and later with no parallel
data at all~\cite{conneau2017word}. In our setting, the analogy is not
a single source-to-target map, but the fact that independently trained
representations can expose compatible coordinates when their inputs are
different views of the same underlying structure. Pulse and motion are
different views of the same physiology, so the practical consequence is
similar: the cross-modal alignment can be estimated from a fraction of
the paired cohort rather than learned from scratch.

\paragraph{Symbols as an interpretable interface: directions and open problems.}
Our results suggest that health FM embeddings
already contain symbol-like structure that can be surfaced
post-hoc.
These symbols are not only interpretable within a single model
but partially shared across modalities, enabling a common
vocabulary through which separately trained models can be
compared and composed.
Connecting symbols to model predictions through causal
circuits~\cite{marks2024sparse} would deepen mechanistic
understanding of what drives health predictions.
Exploring nonlinear extraction methods such as sparse autoencoders
may surface structure missed by linear extractors.
Individual symbol meaning analysis with comprehensive stratified
analysis across subpopulations remains an important open direction.
The symbolic layer also enables targeted modal unlearning, where
specific symbols are suppressed to remove learned biases without
retraining the underlying model.
Symbolic reasoning over extracted components, composing symbols
to derive higher-order health inferences, is a natural next step
toward neurosymbolic health AI.
Finally, projecting symbols from human-uninterpretable modalities
such as accelerometry into more legible spaces such as PPG or ECG
offers a path toward interpretability transfer, where the meaning
of opaque sensor signals is explained in terms of physiologically
familiar ones.

\paragraph{Limitations.}
Our extraction uses linear decompositions, which may miss nonlinear
structure~\cite{engels2024nonlinear}.
We also emphasize that the alignment quality we report is empirical rather
than theoretically guaranteed. Our claims about shared structure describe what these linear operators recover in the data-dependent coordinates they produce on this dataset,
not an extrinsic or canonical correspondence between the embedding spaces.
All three models share the same ${\sim}$172K training population,
so alignment may partly reflect shared covariance; extending this
analysis to models trained on entirely disjoint populations would
further corroborate whether alignment reflects shared physiology
rather than shared demographics.
Three observations argue against a purely demographic explanation:
the blood-type negative control produces no signal despite sharing
population statistics; health conditions transfer with varying
strength (cardiovascular strong, hepatorenal weak); and every
condition exceeds a confounder baseline trained on age, sex, and BMI.

\section{Conclusion}
We showed that frozen health FM embeddings can be
re-expressed via linear decompositions into symbol spaces that are
selectively grounded in health conditions, partially shared across
modalities, and useful for cross-modal transfer with limited paired
data.

\section*{Acknowledgments}
We would like to thank participants in the Apple Heart and Movement
Study and study staff at The Brigham and Women's Hospital, without whom
this work would not have been possible. We are grateful to Vincent Man,
Nandita Bhaskhar, and Lin Yang for valuable internal reviews, and to
Calum MacRae and Angella Spillane for helpful feedback on the
manuscript. Finally, we thank Ian Shapiro, Lindsay Hislop, Laura
Rhodes, and Jonathan Varbel for publication coordination.

\bibliographystyle{icml2026}
\bibliography{references}

\FloatBarrier
\clearpage
\onecolumn
\section{Supplementary}
\vspace{0.1in}

\subsection{Symbol Alignment Details}
\label{sec:alignment_details}

Applying bijective matching to raw embedding dimensions recovers
substantial cross-modal correspondence
(Head-10~$\approx 0.56$, area under the correlation
curve~$\approx 0.32$), establishing a strong baseline.
ICA yields the strongest top-ranked matches but decays rapidly,
suggesting it isolates a small set of highly shared modes.
In contrast, NMF maintains strong alignment across a much broader
range of symbols (area under the correlation curve: 0.39 vs.\ 0.21
for ICA; Figure~\ref{fig:symbol_alignment}b), indicating a more
distributed shared vocabulary.
This behavior is consistent with their inductive biases: ICA favors
statistically independent sources~\cite{hyvarinen2000independent},
whereas NMF's non-negativity promotes parts-based factors that align
more consistently across many
components~\cite{lee1999nmf}.

\subsection{Symbol Meaning Quantification}
\label{sec:symbol_meaning_quantification}

Letting $z$ denote a symbol activation and $y \in \{0,1\}$
indicate condition presence, we compute the Wasserstein-1 (W1)
distance between $p(z \mid y{=}1)$ and $p(z \mid y{=}0)$.
Embeddings are z-scored to unit variance per symbol before
computing W1, so that values are in units of standard deviations.

We use two complementary metrics.
W1 captures full distributional differences (shifts, spreads,
and shape changes), making it sensitive to any systematic
difference between groups. However, W1 is influenced by sample
size imbalance, producing a nonzero floor for minority groups
even under the null.
Cohen's $d$ measures the standardized mean difference and has
$\mathbb{E}[d] = 0$ under the null regardless of group size,
eliminating this floor. However, it only captures mean shifts
and misses higher-order distributional differences.
Together, they complement each other: W1 provides a complete
distributional view while Cohen's $d$ confirms that observed
selectivity reflects genuine effect sizes rather than
sample-size artifacts
(Figure~\ref{fig:cohens_d_si}).

\begin{figure}[t]
\begin{center}
\centerline{\includegraphics[width=\textwidth]{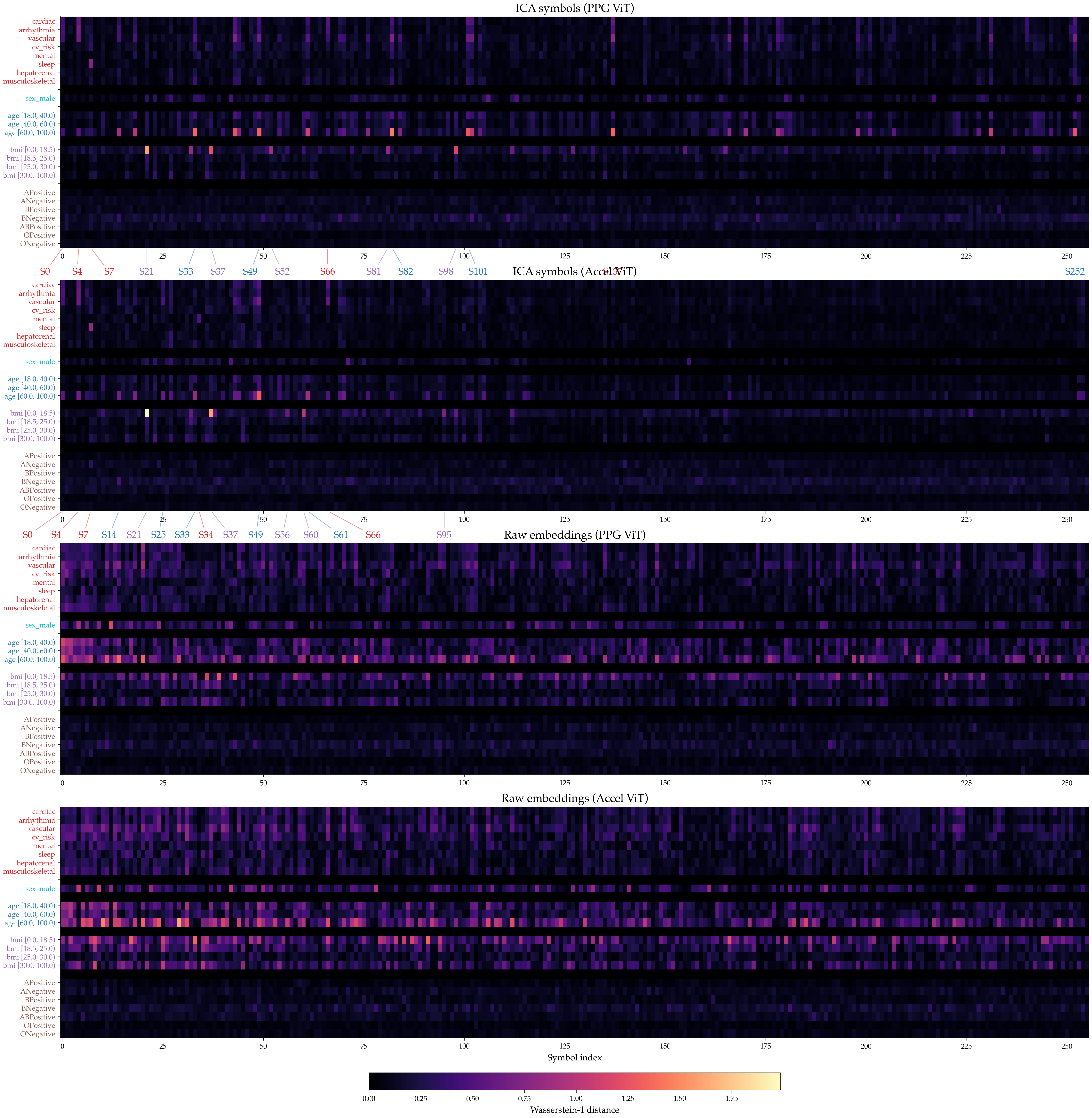}}
\caption{Full symbol-meaning heatmaps (companion to
Figure~\ref{fig:symbol_meaning}a). All 256 symbols are shown for ICA (top two
rows) and raw embeddings (bottom two rows), for PPG~ViT and Accel~ViT, including
the blood-type negative control. Each column is a symbol, each row a target, and
each cell the W1 distance between target-positive and target-negative activation
distributions (shared global color scale). ICA concentrates target associations
into a sparse set of symbols, whereas raw embedding dimensions are diffuse; blood
type produces little signal in any representation. Colored markers indicate the
five highest-SI symbols per target group.}
\label{fig:supp_full_heatmap}
\end{center}
\end{figure}

\subsection{Selectivity Index Details}
\label{sec:si_details}

Both the \emph{shape} (how peaked the profile is)
and the \emph{magnitude} (how strong the associations are) matter
for assessing symbol quality:
a symbol with profile $[0.002, 0.001, 0.001]$ is peaked but
indistinguishable from noise. Standard sparsity measures such as the
Hoyer index~\cite{hoyer2004nonnegative} capture only shape, not
magnitude. The Selectivity Index
(Eq.~\ref{eq:selectivity_index}) combines both.
L1 is preferred over L2 because L2 upweights large values via
squaring, which is redundant with Hoyer's own sensitivity to
concentrated peaks. The product $\mathrm{Hoyer} \times \|\mathbf{w}\|_1$
gives a clean factorization: Hoyer captures shape (how concentrated
the profile is) and L1 captures magnitude (how large the
associations are).

\subsection{Choice of AUC as Transfer Metric}
\label{sec:auc_justification}
Because symbol alignment optimizes correlation rather than scale,
the predicted distribution may shift relative to the source domain.
AUC is invariant to monotonic shifts in the predicted scores. It
depends only on the ranking of predictions, not their absolute
values. This makes AUC a suitable metric for evaluating whether
alignment preserves the discriminative structure of the source
classifier, independent of any scale or offset mismatch introduced
by the alignment operators.

\subsection{Cross-Modal Meaning Consistency}

\begin{figure}[h!]
\begin{center}
\centerline{\includegraphics[width=\columnwidth]{%
  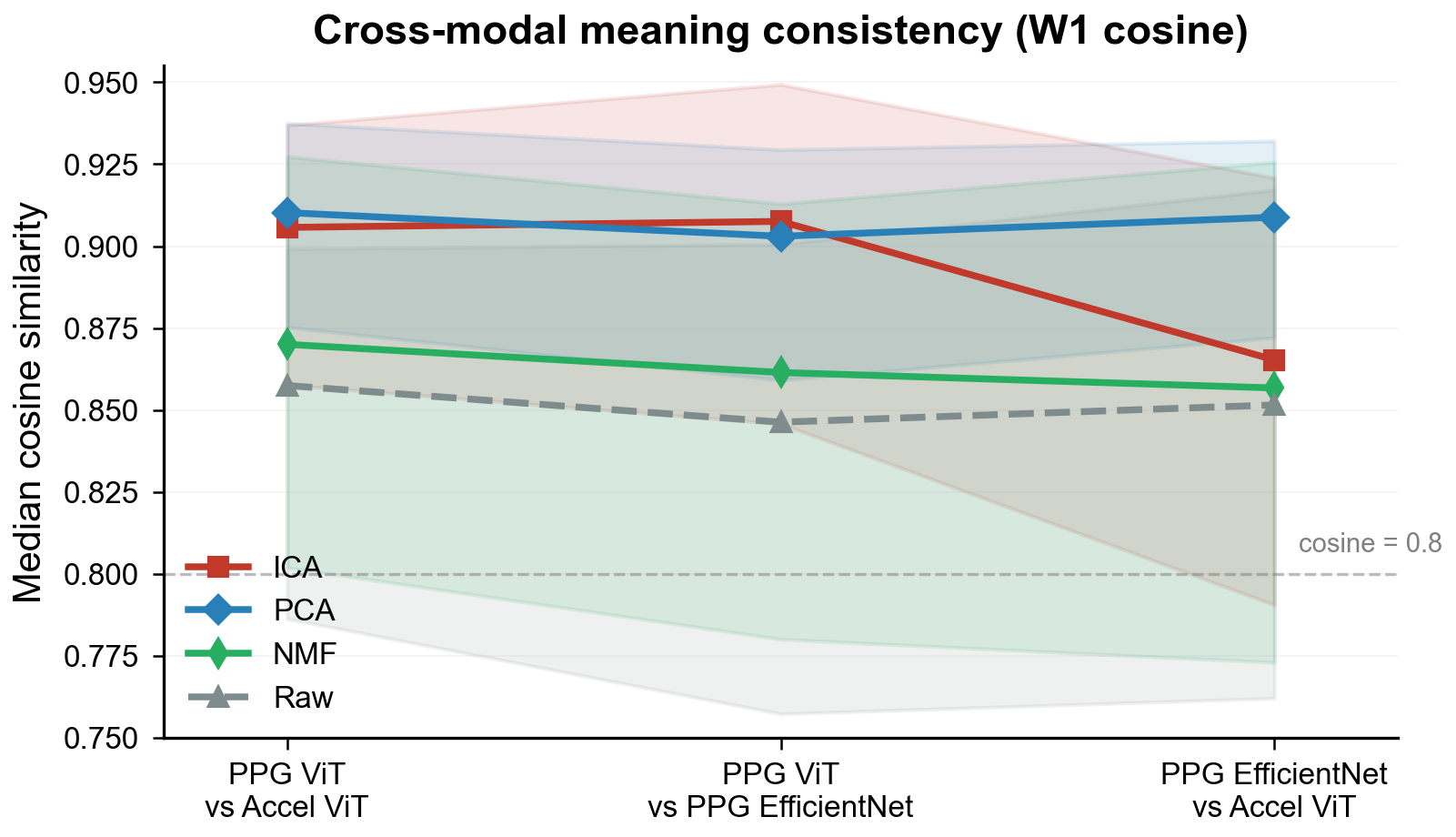}}
\caption{Cross-modal meaning consistency across model pairs and symbol
extraction methods.
Lines show median per-symbol W1 cosine similarity; shaded bands show
interquartile range (Q25--Q75).
PCA maintains 0.90--0.91 across all pairs (architecture-invariant),
while ICA drops from 0.91 to 0.87 when both modality and architecture
differ---higher-order statistics exploited by ICA are sensitive to
architectural nonlinearities.
Even raw embeddings achieve median~0.85, confirming that cross-modal
physiological structure is genuine; symbol extraction adds
disentanglement, sharpening the alignment.}
\label{fig:cross_modal_consistency}
\end{center}
\vskip -0.2in
\end{figure}

Figure~\ref{fig:cross_modal_consistency} shows per-symbol W1 cosine
similarity across three model pairs and four extraction methods.

\subsection{Robustness Check: Cohen's d Selectivity Index}

\begin{figure}[h!]
\begin{center}
\centerline{\includegraphics[width=\textwidth]{%
  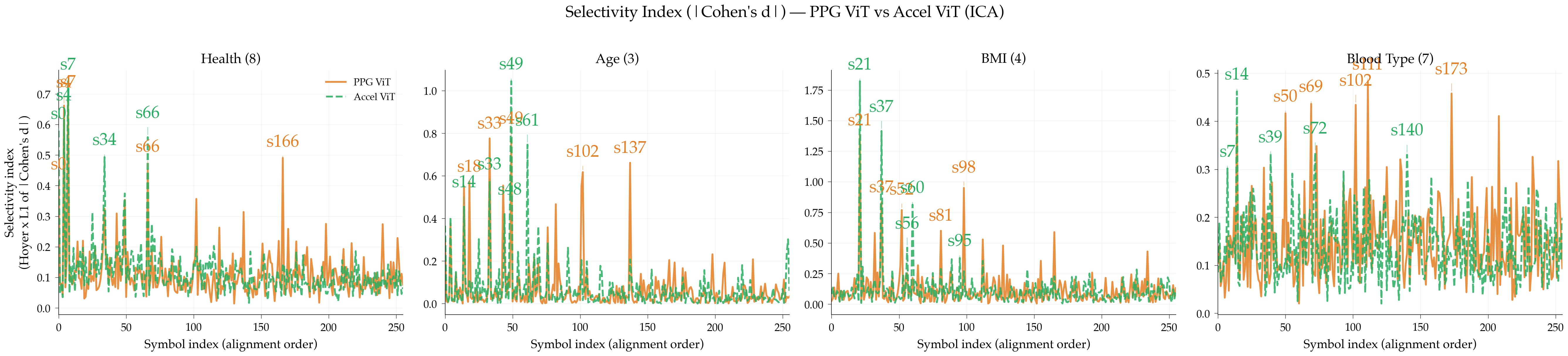}}
\caption{Selectivity Index computed on $|\text{Cohen's } d|$ effect
size (PPG~ViT vs Accel~ViT, ICA+bijective).
Cohen's $d$ has $\mathbb{E}[d] = 0$ under the null regardless of
group size, eliminating the minority-group inflation floor present
in W1.
Age~60+ and BMI underweight remain dominant, confirming genuine
physiological effect sizes.}
\label{fig:cohens_d_si}
\end{center}
\vskip -0.2in
\end{figure}

Figure~\ref{fig:cohens_d_si} replicates the SI analysis using
$|\text{Cohen's } d|$ instead of W1, confirming that the main
findings are robust to the choice of distributional distance.

\subsection{Target Label Definitions}
\label{sec:label_definitions}

Table~\ref{tab:label_definitions} defines the health condition groups used as targets throughout the paper.

\begin{table*}[h!]
\centering
\caption{Health condition group definitions. A subject is labeled
positive if \emph{any} constituent condition is self-reported as
present; negative if \emph{all} are reported absent; otherwise
missing.}\label{tab:label_definitions}
\small
\begin{tabular}{@{}p{3.5cm}p{12.5cm}@{}}
\toprule
\textbf{Group} & \textbf{Positive if any of} \\
\midrule
Cardiac &
  heart attack, heart disease, heart failure,
  beta-blocker use \\[3pt]
Arrhythmia &
  atrial fibrillation, heart rhythm disorder,
  pacemaker \\[3pt]
Vascular &
  stroke/TIA, artery disease, anticoagulant use,
  antiplatelet use, calcium-channel blocker use \\[3pt]
CV risk &
  diabetes, hypertension \\[3pt]
Mental health &
  depression, anxiety, antianxiety use,
  antidepressant use, antipsychotic use \\[3pt]
Sleep &
  sleep apnea, sleep medication use \\[3pt]
Hepatorenal &
  kidney disease, liver disease, urinary disorder,
  diuretic use \\[3pt]
Musculoskeletal &
  arthritis, hip/knee disorder, osteoporosis,
  low-back disorder, neck disorder \\
\midrule
\multicolumn{2}{@{}l}{\textit{Physiological markers}} \\
Sex & binary (male/female) from enrollment \\[3pt]
Age & binned: [18,40), [40,60), [60,100) \\[3pt]
BMI & binned: $<$18.5, 18.5--25, 25--30, 30+ \\
\midrule
\multicolumn{2}{@{}l}{\textit{Negative control}} \\
Blood type &
  ABO/Rh from Apple Health profile;
  one-hot encoded into 7 types \\
\bottomrule
\end{tabular}
\end{table*}

\clearpage
\subsection{Symbol Extraction Methods}
\label{sec:extraction_methods}

Table~\ref{tab:symbol_extractors} summarizes the symbol extraction methods used throughout the paper.

\begin{table}[H]
\centering
\input{tables/table_symbol_extractors}
\end{table}

The methods differ in how the coefficient matrix $Z_m$ is obtained.
For PCA and ICA, once the decomposition is fitted, activations are
computed by an explicit linear readout,
\begin{equation}
Z_m = H_m A_m,
\end{equation}
where $A_m \in \mathbb{R}^{d \times k}$ is the encoder/readout matrix.
For orthonormal PCA, $A_m = S_m$. For ICA, the readout matrix and
reconstruction dictionary are generally distinct, corresponding to the
unmixing and mixing matrices. For NMF, activations are constrained to be
non-negative: $H_m \approx Z_m S_m^\top$ with $Z_m, S_m \ge 0$. They are
therefore estimated by non-negative least-squares fitting against the
learned dictionary rather than by an unconstrained linear readout.

The framework also supports \emph{nonlinear} extractors. A sparse
autoencoder, for example, computes symbol activations through a nonlinear
function,
\begin{equation}
z = \phi_\theta(h), \qquad \hat{h} = \psi_\theta(z).
\end{equation}
Because $\phi_\theta$ and $\psi_\theta$ may include nonlinearities and
learned encoder--decoder structure, the resulting symbols are not
restricted to fixed linear directions in embedding space. Such methods
can in principle capture curved or conditional structure that a single
linear direction would miss, and can also support overcomplete symbol
dictionaries. This is why nonlinear extractors are a natural extension
of the present framework, while the current paper focuses on the
linearly recoverable structure exposed by PCA, ICA, NMF, and CCA-based
alignment.

\subsection{Transfer AUC by Extraction and Alignment Method}

Tables~\ref{tab:transfer_ica}--\ref{tab:transfer_nmf} report
transfer AUC and retention for each symbol extraction method (ICA,
PCA, NMF) across the full range of alignment strategies, from
bijective matching through sparse CCA at six regularization strengths
to dense CCA. Embed+Bijective is included as a baseline that uses
raw embeddings without symbol extraction.
Several patterns emerge.
First, alignment is necessary: the Embed+Bijective baseline already
recovers substantial transfer, but CCA-based alignment consistently
improves over bijective matching for all three extractors.
Second, the choice of extraction method interacts with the alignment
method: PCA with sparse CCA ($\lambda \leq 10^{-3}$) achieves the
highest retention across most targets, while ICA paired with dense
CCA is competitive but less robust across domain pairs.
NMF generally underperforms ICA and PCA, likely because its
non-negativity constraint is a poor match for the centered embedding
space.
Third, blood type targets (negative control) remain near chance
regardless of SE or SA, confirming that alignment does not create
spurious transfer.

\input{tables/table_transfer_se_sa}

\end{document}

%% file: math_commands.tex
\usepackage{amsmath,amsfonts,bm}

\def\eqref#1{equation~\ref{#1}}
\def\1{\bm{1}}

\DeclareMathAlphabet{\mathsfit}{\encodingdefault}{\sfdefault}{m}{sl}
\SetMathAlphabet{\mathsfit}{bold}{\encodingdefault}{\sfdefault}{bx}{n}

%% file: tables/target_groups_table.tex
% Real-world target groups for symbol meaning validation
% Include in paper with: \input{tables/target_groups_table.tex}

\begin{table}[t]
\centering
\caption{Target condition groups for symbol meaning
analysis.}\label{tab:target_groups}
\small
\begin{tabular}{@{}p{1.8cm}p{3.6cm}cc@{}}
\toprule
\textbf{Group} & \textbf{Labels} & $n$ & \textbf{Role} \\
\midrule
Health conditions &
  cardiac, arrhythmia, vascular, cv~risk, mental, sleep,
  hepatorenal, musculoskeletal
  & 8 & Positive \\[4pt]
Physiological markers &
  sex~(male), age~(18--40, 40--60, 60+),
  BMI~($<$18.5, 18.5--25, 25--30, 30+)
  & 8 & Positive \\[4pt]
\midrule
Blood type &
  A+, A$-$, B+, B$-$, AB+, O+, O$-$
  & 7 & Negative \\[2pt]
\bottomrule
\end{tabular}
\vspace{4pt}

\raggedright
\footnotesize
\emph{Positive}: health conditions and physiological markers
affect PPG waveform morphology and motion patterns through
known physiological pathways.
\emph{Negative}: ABO/Rh antigens have no proximal mechanism
to affect PPG or motion signals.
Total: 23 conditions across 3 groups.
\end{table}

%% file: results_symbol_alignment.tex
\subsection{Do Symbols Align Across Modalities?}
\label{sec:symbol_alignment}

\begin{figure}[t]
\begin{center}
\centerline{\includegraphics[width=\columnwidth]{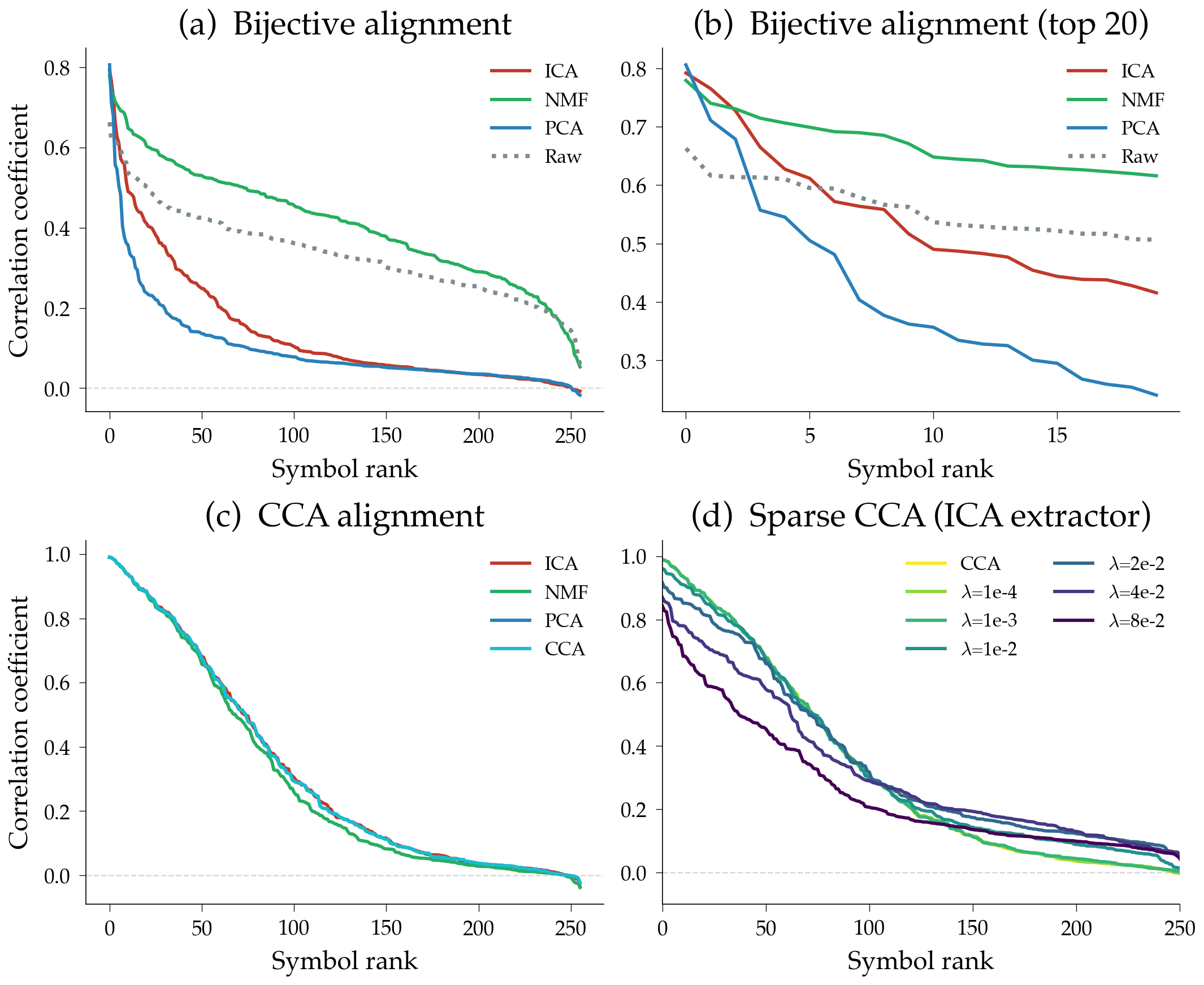}}
\caption{Cross-modal symbol alignment analysis [PPG~EfficientNet vs Accel~ViT].
\textbf{(a)}~Bijective matching across symbol extractors: cross-modal
correspondence is already present in raw embeddings (black dotted).
Symbol extraction reshapes how it is distributed --- ICA concentrates
it into a few top-ranked symbols, while NMF spreads it more broadly
across ranks.
\textbf{(b)}~Zoomed view of top-20 symbols showing ICA's stronger
peak versus NMF's consistently high alignment.
\textbf{(c)}~CCA alignment: all extractors converge to near-identical
curves, indicating strongly shared cross-modal linear
structure that is less sensitive to extraction basis.
\textbf{(d)}~Sparse CCA: L1 penalty on projection weights
$\leq 2 \times 10^{-2}$ preserves alignment while yielding sparser
projections suggesting that
much of the shared linear structure can be described by relatively
few dimensions.}
\label{fig:symbol_alignment}
\end{center}
\end{figure}

If independently trained models on different sensor modalities converge on
a shared representational structure, their learned features should align
across modalities.
Figure~\ref{fig:symbol_alignment} shows results for
PPG~EfficientNet vs Accel~ViT, the hardest pair (different modality
and architecture); the other two pairs show similar patterns.

\paragraph{Cross-modal correspondence is present in embeddings and symbol extraction reshapes it.}
Comparing raw embedding dimensions without alignment yields
near-zero cross-modal correlation.
Bijective alignment recovers substantial correspondence even on
raw embeddings (Figure~\ref{fig:symbol_alignment}a), and symbol
extraction further reshapes how this correspondence is distributed
across components. Overall, cross-modal correspondence is already
present in the raw embeddings; symbol extraction reshapes how it
is distributed, with ICA concentrating it into a few top symbols
and NMF spreading it more broadly
(see Supplementary~\ref{sec:alignment_details} for details).

\paragraph{CCA reveals shared cross-modal linear structure.}
Differences between extractors largely disappeared under CCA alignment
(Figure~\ref{fig:symbol_alignment}c).
All representations converged to near-identical performance
($r \approx 0.99$ at top ranks).
This convergence is expected because CCA~\cite{hotelling1936cca} solves
for the linear projections that maximize cross-modal correlation,
effectively bypassing whatever basis the extractor produced.
These results suggest that PPG and accelerometer embeddings
share a linearly accessible subspace, and that CCA recovers a maximally correlated coordinate system
for it. Whether this subspace encodes meaningful physiology is
tested in Section~\ref{sec:symbol_meaning}.

\paragraph{Sparse CCA preserves alignment with interpretability.}
As shown above, CCA alignment is insensitive to the choice of
extractor, but its canonical projections are dense and individual
symbols difficult to interpret.
We examined whether interpretability could be improved by imposing
sparsity on the CCA projection weights without sacrificing alignment
(Figure~\ref{fig:symbol_alignment}d).
Low-to-moderate L1 regularization
($\lambda \leq 2 \times 10^{-2}$) maintained $>95\%$ of standard CCA's
alignment capacity while yielding sparser projections.
Only at high sparsity ($\lambda > 4 \times 10^{-2}$) did alignment
degrade substantially. This suggests that much of the shared linear structure can be
captured by a relatively small number of contributing dimensions.

%% file: results_symbol_meaning_analysis.tex
\subsection{Do Symbols Carry Interpretable Health Meaning?}
\label{sec:symbol_meaning}

\begin{figure*}[t]
\vspace{-0.3in}
\begin{center}
\centerline{\includegraphics[width=\textwidth,keepaspectratio]{%
  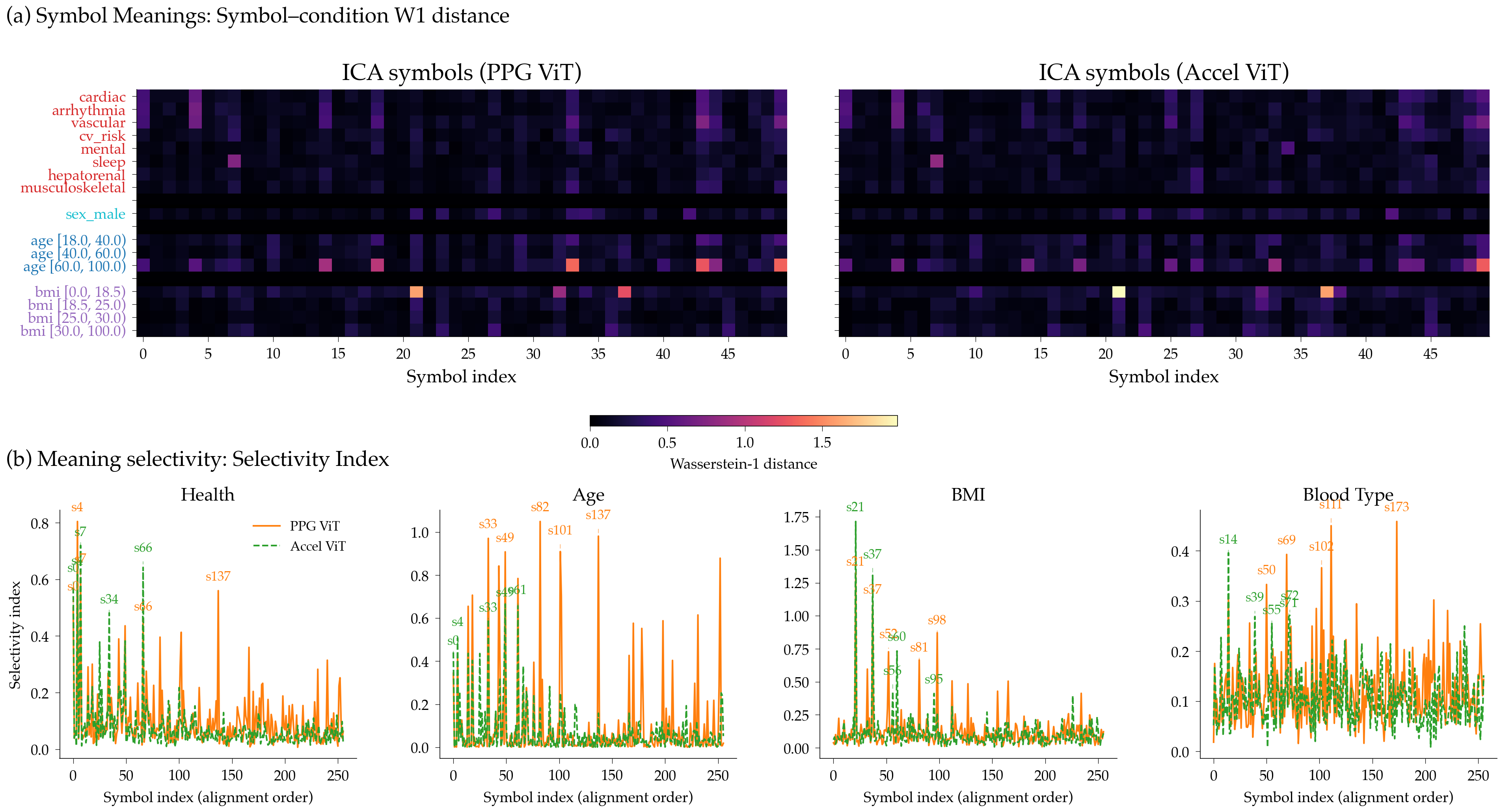}}
\caption{Symbol meaning and selectivity (ICA, PPG~ViT vs Accel~ViT).
\textbf{(a)}~W1 heatmaps for the 50 most cross-modally aligned ICA symbols:
each column is a symbol (an extracted direction in embedding space), each row a
health or physiological target, and each cell the W1 distance between the
symbol's activation distributions for target-positive vs target-negative
subjects (higher values indicate a stronger association). PPG~ViT (left) and
Accel~ViT (right) exhibit the same selective structure --- most cells are near
zero while a few symbols activate for specific targets. The full set of 256
symbols, the raw-embedding comparison, and the blood-type control are shown in
Supplementary Figure~\ref{fig:supp_full_heatmap}.
\textbf{(b)}~Selectivity Index (SI~$= H \cdot \|\mathbf{w}\|_1$) per symbol in
alignment order for each target group; the five highest-SI symbols per curve are
labeled. Health conditions, age, and BMI reach high SI, whereas blood type
(negative control) remains uniformly low in both modalities.}
\label{fig:symbol_meaning}
\end{center}
\vskip -0.2in
\end{figure*}

Having established that symbols align across modalities,
we next ask whether they carry selective, interpretable meaning.
We focus on PPG~ViT vs Accel~ViT
(Figure~\ref{fig:symbol_meaning}), which share the same
architecture so that differences in symbol meaning reflect
modality rather than model design; other model pairs show similar
patterns (Supplementary Figure~\ref{fig:cross_modal_consistency}).

\paragraph{Symbols show semantic grounding and selective meaning.}
We focus on ICA because its independence constraint produces the
sparsest and most interpretable target profiles.
Figure~\ref{fig:symbol_meaning}a shows W1 heatmaps for PPG~ViT
and Accel~ViT,
where each cell shows the association strength between the
corresponding symbol (column) and target (row).
 Individual symbols activate for specific
targets while most symbol-target pairs are near zero, quantified
per symbol by the Selectivity Index (Eq.~\ref{eq:selectivity_index}).
The same selective structure appears in both modalities,
confirming that the underlying models encode shared health-relevant
structure. The corresponding raw-embedding heatmaps, which are far
more diffuse, and the blood-type negative control, which produces
little or no signal in any representation, are shown in
Supplementary Figure~\ref{fig:supp_full_heatmap}.

\textit{While detailed interpretation of individual symbols is beyond
the scope of this paper}, a few patterns are worth noting.
Symbol S7 peaks at sleep and is the brightest accelerometer symbol,
consistent with sleep patterns directly affecting motion. In the
PPG heatmap, S7 does not overlap with other health targets, whereas
heart-related conditions (cardiac, vascular, arrhythmia) share
clustered symbols, reflecting common cardiovascular physiology.
Hepatorenal and mental health targets show weaker cross-modal
correspondence, consistent with more indirect physiological
signatures in wearable data.

Figure~\ref{fig:symbol_meaning}b quantifies this selectivity using
the Selectivity Index (SI, Eq.~\ref{eq:selectivity_index}) per
symbol in alignment order.
Accelerometer SI peaks concentrate among the most cross-modally
aligned symbols, whereas PPG distributes SI more broadly, consistent
with the richer physiological encoding of the pulse waveform.
Blood type SI remains uniformly low in both modalities and,
unlike health conditions, shows no cross-modal consistency,
confirming that the negative control carries neither selective
signal nor shared structure.

\paragraph{Quantifying cross-modal meaning consistency.}
Having seen this consistency qualitatively in the heatmaps and
SI concentration (Figure~\ref{fig:symbol_meaning}a,b), we measure it
directly as the cosine similarity between each symbol's W1 target
profile in PPG and Accel. Even raw embeddings reach a median cosine
of 0.84, so the cross-modal structure is already present, and ICA
sharpens it further, with 91\% of symbols above cosine 0.8 under
bijective alignment (median 0.89). Supplementary
Figure~\ref{fig:cross_modal_consistency} extends this across three
model pairs and four extraction methods with broadly consistent
results.

The selectivity results above are robust to replacing W1 with $|\text{Cohen's } d|$ as the group-difference metric (Supplementary Figure~\ref{fig:cohens_d_si}).

%% file: results_symbolic_transfer.tex
\subsection{Can Symbols Support Cross-Domain Transfer?}
\label{sec:symbolic_transfer}

We evaluated cross-modal symbolic transfer using all three foundation
models and six directed domain pairs using the pipeline described in
\S\ref{sec:symbolic_transfer_method}.
\begin{figure*}[!htbp]
\begin{center}
\centerline{\includegraphics[width=\textwidth]{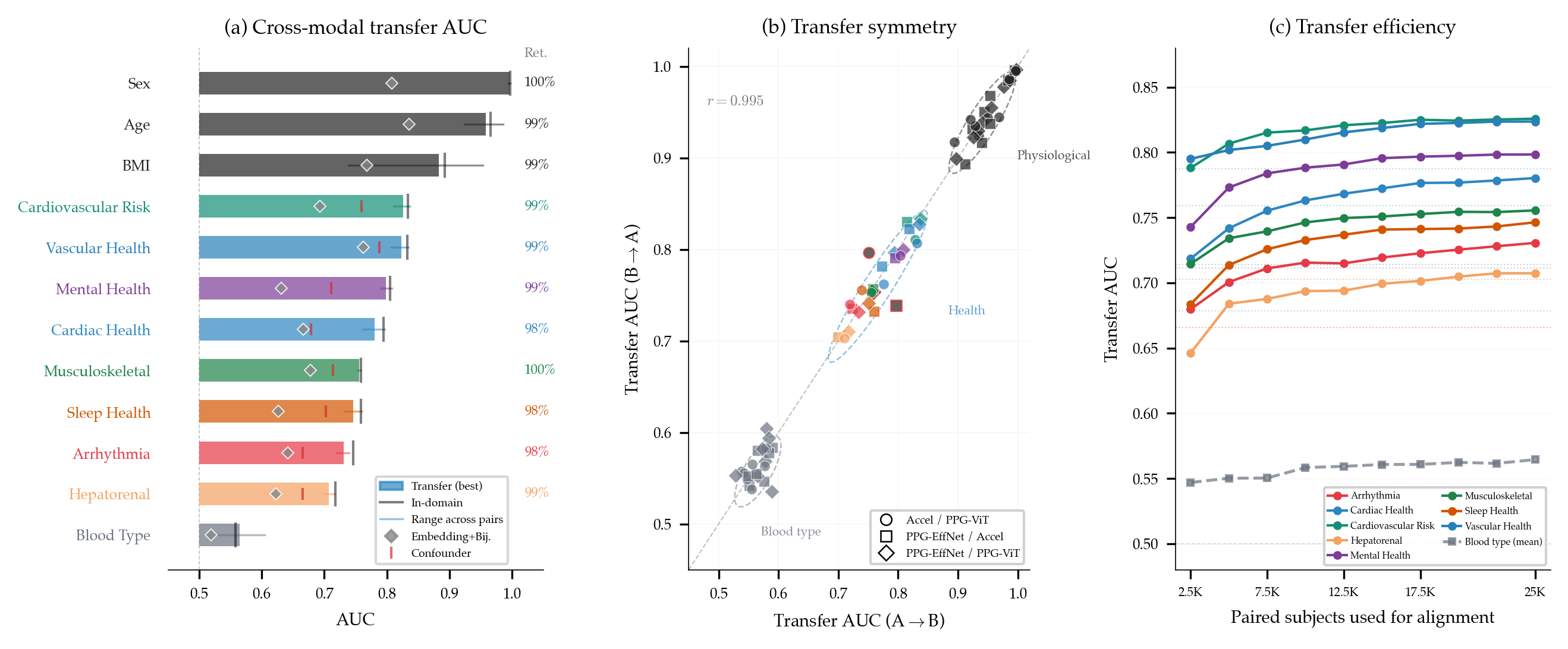}}
\caption{Cross-modal symbolic transfer.
\textbf{(a)}~Classifiers trained in one symbol space retain $>$98\% of
in-domain AUC after cross-modal alignment; blood type
(negative control) remains near chance. Best transfer is
achieved with the CCA family of alignments.
\textbf{(b)}~Transfer retention is similar in both directions
($r = 0.995$), consistent with a low-complexity alignment recoverable by linear maps; we report this empirically and do not claim a formal isomorphism between symbol spaces.
\textbf{(c)}~Transfer saturates with a fraction of the paired cohort, suggesting
a low-complexity alignment.
Overall, independently trained health models share
physiologically rich structure that can be
leveraged for cross-modal transfer without joint training.}
\label{fig:symbolic_transfer}
\end{center}
\vskip -0.2in
\end{figure*}

\paragraph{Symbols enable cross-modal transfer without joint training.}
Figure~\ref{fig:symbolic_transfer}a summarizes transfer
performance; best results are achieved with the CCA family of alignments.
Physiological markers transfer most readily (AUC 0.88--1.00,
$\approx$100\% retention). Health conditions follow with AUC
0.71--0.83 ($98.7\% \pm 1.6\%$ retention), while blood type
remains near chance (AUC $\approx 0.56$) as expected for a
negative control (full per-target AUC in Supplementary
Tables~\ref{tab:transfer_ica}--\ref{tab:transfer_nmf}). Without symbol extraction and alignment, raw embedding
transfer yields near-chance AUC ($\sim$0.50). Bijective
matching on raw embeddings (grey diamonds in
Figure~\ref{fig:symbolic_transfer}a) recovers partial transfer (health
mean AUC $= 0.67$) but remains well below CCA, because raw dimensions are entangled and a single bijective
mapping cannot disentangle them.

A natural concern is that cross-modal transfer merely propagates
demographic information shared between modalities. To rule this out, we
include a per-condition confounder baseline that predicts each health
target from age, sex, and BMI alone (red ticks in
Figure~\ref{fig:symbolic_transfer}a). Every health condition
exceeds its confounder baseline, confirming that symbolic transfer
carries condition-specific information that demographics alone do not
explain.

\paragraph{Transfer retention is similar in both directions.}
Figure~\ref{fig:symbolic_transfer}b shows that forward and reverse
transfer AUC are tightly correlated across all targets and domain
pairs ($r = 0.995$, mean $|\Delta\text{AUC}| = 0.009$).
The high retention above, together with its similarity across both
transfer directions, indicates that the two representations carry much of
the same predictive information with no direction-specific loss. How much paired data is needed to
estimate the cross-modal alignment map (the linear map that places the
two symbol spaces in correspondence) is a separate question that we
examine next.

\paragraph{Transfer retention plateaus before the full paired cohort.}
We evaluate how much paired data is needed for effective
cross-modal transfer by sweeping the number of paired subjects
used for alignment (Figure~\ref{fig:symbolic_transfer}c). All
eight health conditions plateau above 95\% retention by
$\sim$7.5K paired subjects, roughly 30\% of the ${\sim}$24K
operator-fit cohort, with only marginal gains from
additional data. We discuss the implications of this
saturation in \S\ref{sec:discussion}.

%% file: tables/table_symbol_extractors.tex
% \begin{table}[t]
\centering
\caption{Unified view of the symbol extraction methods used to
re-express frozen embeddings in alternative coordinate systems.
Each method is an extraction operator $\mathcal{E}$ mapping embeddings
$X$ to symbol activations $Z$, with a corresponding reconstruction back
to the embedding space. For PCA and ICA the activations are an explicit
linear readout $Z = X A$ ($A$ the readout/encoder matrix); NMF has no
such readout and obtains $Z$ by non-negative least squares against the
dictionary $S$. The dictionary $S$ (columns are the candidate symbol
directions) reconstructs embeddings via $\hat{X} = Z S^\top$; for
orthonormal PCA the readout and dictionary coincide ($A = S$). In
nonlinear cases, reconstruction is a learned inverse without assuming a
linear symbol basis. We use the activations $Z$ directly; the
reconstruction column is included only to characterize each method.}
\label{tab:symbol_extractors}
\begin{tabular}{p{2.6cm} p{3.4cm} p{3.4cm} p{3.8cm}}
\toprule
\textbf{Method} &
\textbf{Primary Assumption / Inductive Bias} &
\textbf{Extraction} &
\textbf{Reconstruction} \\
\midrule

PCA &
Dominant structure captured by maximal variance; orthogonality &
$Z = X A$ &
$\hat{X} = Z S^\top \;\approx X$ \\

ICA &
Latent factors are statistically independent &
$Z = X A$ &
$\hat{X} = Z S^\top \;\approx X$ \\

NMF &
Representations are additive and non-negative &
$Z = \arg\min_{Z \ge 0} \lVert X - Z S^\top\rVert^2,\; S \ge 0$ &
$\hat{X} = Z S^\top$ \\

CCA &
Shared structure exists between paired embedding spaces &
$Z_1 = X_1 A_1,\; Z_2 = X_2 A_2$ &
$\hat{X}_1 \approx Z_1 S_1^\top,\;
 \hat{X}_2 \approx Z_2 S_2^\top$ \\

Sparse Autoencoder &
Meaningful structure expressed via sparse, reusable activations &
$Z = \phi_\theta(X)$ &
$\hat{X} = \psi_\theta(Z) \;\approx X$ \\

\bottomrule
\end{tabular}
% \end{table}

%% file: tables/table_transfer_se_sa.tex
\begin{table*}[h!]
\centering
\caption{Cross-modal transfer AUC and retention (mean across six domain pairs) for ICA symbol extraction across alignment methods. Each cell shows transfer AUC with retention in parentheses. Sparse CCA columns show decreasing $\lambda$.}
\label{tab:transfer_ica}
\small
\setlength{\tabcolsep}{3pt}
\begin{tabular}{@{}lc c cccccccc @{}}
\toprule
 & & Embed & \multicolumn{8}{c}{ICA} \\
\cmidrule(lr){3-3} \cmidrule(lr){4-11}
\textbf{Target} & In-d. & Bij. & Bij. & $8{\times}10^{-2}$ & $4{\times}10^{-2}$ & $2{\times}10^{-2}$ & $10^{-2}$ & $10^{-3}$ & $10^{-4}$ & CCA \\
\midrule
Cardiac & 0.80 & 0.67 (84\%) & 0.68 (86\%) & 0.57 (74\%) & 0.62 (79\%) & 0.68 (85\%) & 0.64 (81\%) & 0.68 (86\%) & 0.69 (87\%) & 0.69 (87\%) \\
Arrhythmia & 0.75 & 0.64 (86\%) & 0.67 (92\%) & 0.60 (82\%) & 0.64 (87\%) & 0.64 (88\%) & 0.65 (88\%) & 0.65 (89\%) & 0.65 (88\%) & 0.65 (89\%) \\
Vascular & 0.83 & 0.76 (92\%) & 0.74 (89\%) & 0.59 (72\%) & 0.67 (81\%) & 0.68 (82\%) & 0.71 (86\%) & 0.71 (85\%) & 0.73 (87\%) & 0.73 (87\%) \\
CV Risk & 0.83 & 0.69 (83\%) & 0.70 (84\%) & 0.56 (69\%) & 0.62 (74\%) & 0.68 (82\%) & 0.72 (86\%) & 0.72 (86\%) & 0.72 (86\%) & 0.72 (87\%) \\
Mental & 0.81 & 0.63 (78\%) & 0.69 (85\%) & 0.57 (72\%) & 0.60 (74\%) & 0.65 (81\%) & 0.68 (85\%) & 0.70 (87\%) & 0.71 (89\%) & 0.71 (88\%) \\
Sleep & 0.76 & 0.63 (83\%) & 0.68 (90\%) & 0.60 (80\%) & 0.61 (81\%) & 0.65 (86\%) & 0.63 (84\%) & 0.67 (88\%) & 0.66 (88\%) & 0.66 (88\%) \\
Hepatorenal & 0.72 & 0.62 (87\%) & 0.63 (89\%) & 0.54 (77\%) & 0.60 (85\%) & 0.63 (88\%) & 0.63 (89\%) & 0.61 (86\%) & 0.64 (90\%) & 0.64 (90\%) \\
Musculoskeletal & 0.76 & 0.68 (89\%) & 0.67 (88\%) & 0.53 (71\%) & 0.61 (81\%) & 0.64 (85\%) & 0.66 (87\%) & 0.67 (89\%) & 0.67 (89\%) & 0.67 (89\%) \\
\midrule
Sex & 1.00 & 0.81 (81\%) & 0.83 (84\%) & 0.66 (66\%) & 0.75 (75\%) & 0.80 (80\%) & 0.82 (82\%) & 0.82 (82\%) & 0.82 (82\%) & 0.81 (82\%) \\
Age 18--39 & 0.96 & 0.86 (90\%) & 0.77 (81\%) & 0.56 (59\%) & 0.64 (67\%) & 0.74 (77\%) & 0.77 (81\%) & 0.81 (85\%) & 0.82 (86\%) & 0.81 (85\%) \\
Age 40--59 & 0.94 & 0.78 (83\%) & 0.74 (78\%) & 0.58 (63\%) & 0.60 (64\%) & 0.75 (79\%) & 0.75 (80\%) & 0.80 (85\%) & 0.80 (85\%) & 0.80 (85\%) \\
Age 60+ & 0.99 & 0.89 (90\%) & 0.88 (89\%) & 0.70 (72\%) & 0.75 (76\%) & 0.80 (81\%) & 0.82 (84\%) & 0.85 (87\%) & 0.85 (86\%) & 0.85 (86\%) \\
\midrule
BMI $<$18.5 & 0.95 & 0.85 (90\%) & 0.82 (87\%) & 0.78 (83\%) & 0.76 (81\%) & 0.80 (85\%) & 0.79 (84\%) & 0.80 (85\%) & 0.81 (86\%) & 0.81 (86\%) \\
BMI 18.5--25 & 0.91 & 0.81 (89\%) & 0.72 (79\%) & 0.59 (66\%) & 0.63 (70\%) & 0.71 (79\%) & 0.76 (84\%) & 0.77 (84\%) & 0.77 (85\%) & 0.78 (86\%) \\
BMI 25--30 & 0.78 & 0.57 (73\%) & 0.65 (84\%) & 0.54 (71\%) & 0.55 (71\%) & 0.61 (79\%) & 0.65 (83\%) & 0.67 (86\%) & 0.68 (88\%) & 0.68 (88\%) \\
BMI 30+ & 0.93 & 0.84 (90\%) & 0.76 (81\%) & 0.60 (65\%) & 0.66 (70\%) & 0.76 (81\%) & 0.78 (84\%) & 0.78 (84\%) & 0.79 (84\%) & 0.79 (85\%) \\
\midrule
A+ & 0.56 & 0.53 (94\%) & 0.52 (94\%) & 0.52 (93\%) & 0.52 (93\%) & 0.53 (94\%) & 0.52 (93\%) & 0.52 (94\%) & 0.53 (96\%) & 0.53 (96\%) \\
A$-$ & 0.59 & 0.54 (95\%) & 0.54 (97\%) & 0.53 (96\%) & 0.54 (97\%) & 0.54 (96\%) & 0.53 (96\%) & 0.53 (96\%) & 0.53 (96\%) & 0.53 (95\%) \\
B+ & 0.58 & 0.55 (95\%) & 0.53 (93\%) & 0.51 (90\%) & 0.53 (94\%) & 0.54 (94\%) & 0.53 (94\%) & 0.55 (96\%) & 0.55 (96\%) & 0.55 (96\%) \\
B$-$ & 0.51 & 0.51 (107\%) & 0.50 (105\%) & 0.49 (102\%) & 0.48 (101\%) & 0.50 (107\%) & 0.49 (105\%) & 0.48 (102\%) & 0.47 (99\%) & 0.47 (100\%) \\
AB+ & 0.54 & 0.49 (94\%) & 0.51 (97\%) & 0.53 (101\%) & 0.52 (100\%) & 0.52 (99\%) & 0.53 (101\%) & 0.53 (101\%) & 0.52 (98\%) & 0.52 (98\%) \\
O+ & 0.58 & 0.52 (92\%) & 0.54 (96\%) & 0.53 (94\%) & 0.52 (93\%) & 0.54 (95\%) & 0.53 (94\%) & 0.54 (96\%) & 0.54 (96\%) & 0.54 (96\%) \\
O$-$ & 0.54 & 0.50 (94\%) & 0.51 (96\%) & 0.50 (95\%) & 0.52 (98\%) & 0.52 (98\%) & 0.52 (97\%) & 0.52 (98\%) & 0.52 (97\%) & 0.52 (97\%) \\
\bottomrule
\end{tabular}
\end{table*}

\begin{table*}[h!]
\centering
\caption{Cross-modal transfer AUC and retention (mean across six domain pairs) for PCA symbol extraction across alignment methods. Each cell shows transfer AUC with retention in parentheses. Sparse CCA columns show decreasing $\lambda$.}
\label{tab:transfer_pca}
\small
\setlength{\tabcolsep}{3pt}
\begin{tabular}{@{}lc c cccccccc @{}}
\toprule
 & & Embed & \multicolumn{8}{c}{PCA} \\
\cmidrule(lr){3-3} \cmidrule(lr){4-11}
\textbf{Target} & In-d. & Bij. & Bij. & $8{\times}10^{-2}$ & $4{\times}10^{-2}$ & $2{\times}10^{-2}$ & $10^{-2}$ & $10^{-3}$ & $10^{-4}$ & CCA \\
\midrule
Cardiac & 0.80 & 0.67 (84\%) & 0.55 (69\%) & 0.67 (85\%) & 0.76 (96\%) & 0.74 (94\%) & 0.76 (96\%) & 0.75 (95\%) & 0.78 (98\%) & 0.68 (86\%) \\
Arrhythmia & 0.75 & 0.64 (86\%) & 0.58 (79\%) & 0.69 (93\%) & 0.70 (94\%) & 0.71 (96\%) & 0.72 (97\%) & 0.72 (98\%) & 0.73 (99\%) & 0.64 (87\%) \\
Vascular & 0.83 & 0.76 (92\%) & 0.65 (78\%) & 0.80 (96\%) & 0.80 (96\%) & 0.79 (95\%) & 0.79 (95\%) & 0.79 (95\%) & 0.78 (94\%) & 0.71 (85\%) \\
CV Risk & 0.83 & 0.69 (83\%) & 0.63 (76\%) & 0.71 (86\%) & 0.74 (90\%) & 0.78 (93\%) & 0.79 (95\%) & 0.82 (98\%) & 0.82 (99\%) & 0.71 (85\%) \\
Mental & 0.81 & 0.63 (78\%) & 0.58 (73\%) & 0.72 (89\%) & 0.69 (86\%) & 0.75 (94\%) & 0.76 (94\%) & 0.79 (98\%) & 0.80 (99\%) & 0.70 (87\%) \\
Sleep & 0.76 & 0.63 (83\%) & 0.54 (72\%) & 0.67 (88\%) & 0.68 (90\%) & 0.69 (92\%) & 0.70 (93\%) & 0.74 (98\%) & 0.75 (99\%) & 0.67 (89\%) \\
Hepatorenal & 0.72 & 0.62 (87\%) & 0.53 (74\%) & 0.64 (89\%) & 0.66 (93\%) & 0.68 (95\%) & 0.69 (97\%) & 0.69 (97\%) & 0.71 (99\%) & 0.63 (88\%) \\
Musculoskeletal & 0.76 & 0.68 (89\%) & 0.60 (79\%) & 0.66 (87\%) & 0.69 (91\%) & 0.69 (91\%) & 0.73 (96\%) & 0.74 (98\%) & 0.75 (99\%) & 0.67 (88\%) \\
\midrule
Sex & 1.00 & 0.81 (81\%) & 0.57 (57\%) & 0.96 (97\%) & 0.98 (99\%) & 0.98 (99\%) & 0.99 (99\%) & 1.00 (100\%) & 1.00 (100\%) & 0.83 (84\%) \\
Age 18--39 & 0.96 & 0.86 (90\%) & 0.60 (63\%) & 0.87 (92\%) & 0.87 (91\%) & 0.89 (94\%) & 0.94 (98\%) & 0.95 (99\%) & 0.95 (99\%) & 0.80 (84\%) \\
Age 40--59 & 0.94 & 0.78 (83\%) & 0.57 (60\%) & 0.82 (87\%) & 0.79 (84\%) & 0.87 (93\%) & 0.89 (94\%) & 0.90 (96\%) & 0.93 (99\%) & 0.79 (84\%) \\
Age 60+ & 0.99 & 0.89 (90\%) & 0.61 (62\%) & 0.95 (96\%) & 0.92 (93\%) & 0.96 (97\%) & 0.97 (99\%) & 0.98 (99\%) & 0.98 (100\%) & 0.82 (83\%) \\
\midrule
BMI $<$18.5 & 0.95 & 0.85 (90\%) & 0.57 (61\%) & 0.91 (97\%) & 0.88 (94\%) & 0.93 (98\%) & 0.93 (99\%) & 0.93 (99\%) & 0.93 (99\%) & 0.80 (85\%) \\
BMI 18.5--25 & 0.91 & 0.81 (89\%) & 0.55 (60\%) & 0.84 (93\%) & 0.78 (86\%) & 0.85 (93\%) & 0.86 (95\%) & 0.89 (98\%) & 0.90 (99\%) & 0.77 (85\%) \\
BMI 25--30 & 0.78 & 0.57 (73\%) & 0.52 (67\%) & 0.61 (79\%) & 0.63 (80\%) & 0.68 (88\%) & 0.69 (89\%) & 0.75 (96\%) & 0.76 (98\%) & 0.68 (88\%) \\
BMI 30+ & 0.93 & 0.84 (90\%) & 0.56 (60\%) & 0.85 (91\%) & 0.88 (94\%) & 0.90 (96\%) & 0.90 (97\%) & 0.92 (99\%) & 0.92 (99\%) & 0.79 (84\%) \\
\midrule
A+ & 0.56 & 0.53 (94\%) & 0.51 (92\%) & 0.54 (98\%) & 0.52 (94\%) & 0.52 (94\%) & 0.53 (95\%) & 0.54 (96\%) & 0.55 (98\%) & 0.53 (95\%) \\
A$-$ & 0.59 & 0.54 (95\%) & 0.50 (91\%) & 0.55 (96\%) & 0.53 (94\%) & 0.53 (95\%) & 0.54 (98\%) & 0.53 (96\%) & 0.54 (97\%) & 0.51 (92\%) \\
B+ & 0.58 & 0.55 (95\%) & 0.50 (87\%) & 0.55 (96\%) & 0.55 (96\%) & 0.57 (100\%) & 0.56 (97\%) & 0.58 (101\%) & 0.58 (101\%) & 0.55 (97\%) \\
B$-$ & 0.51 & 0.51 (107\%) & 0.47 (99\%) & 0.50 (104\%) & 0.52 (110\%) & 0.48 (101\%) & 0.47 (102\%) & 0.46 (99\%) & 0.46 (98\%) & 0.48 (101\%) \\
AB+ & 0.54 & 0.49 (94\%) & 0.50 (96\%) & 0.51 (98\%) & 0.52 (99\%) & 0.53 (100\%) & 0.53 (102\%) & 0.53 (101\%) & 0.53 (101\%) & 0.53 (101\%) \\
O+ & 0.58 & 0.52 (92\%) & 0.50 (89\%) & 0.54 (94\%) & 0.54 (96\%) & 0.54 (95\%) & 0.54 (96\%) & 0.55 (98\%) & 0.56 (99\%) & 0.54 (96\%) \\
O$-$ & 0.54 & 0.50 (94\%) & 0.52 (98\%) & 0.51 (95\%) & 0.51 (95\%) & 0.51 (97\%) & 0.51 (97\%) & 0.53 (99\%) & 0.53 (100\%) & 0.52 (98\%) \\
\bottomrule
\end{tabular}
\end{table*}

\begin{table*}[h!]
\centering
\caption{Cross-modal transfer AUC and retention (mean across six domain pairs) for NMF symbol extraction across alignment methods. Each cell shows transfer AUC with retention in parentheses. Sparse CCA columns show decreasing $\lambda$.}
\label{tab:transfer_nmf}
\small
\setlength{\tabcolsep}{3pt}
\begin{tabular}{@{}lc c cccccccc @{}}
\toprule
 & & Embed & \multicolumn{8}{c}{NMF} \\
\cmidrule(lr){3-3} \cmidrule(lr){4-11}
\textbf{Target} & In-d. & Bij. & Bij. & $8{\times}10^{-2}$ & $4{\times}10^{-2}$ & $2{\times}10^{-2}$ & $10^{-2}$ & $10^{-3}$ & $10^{-4}$ & CCA \\
\midrule
Cardiac & 0.80 & 0.67 (84\%) & 0.50 (63\%) & 0.54 (69\%) & 0.63 (80\%) & 0.64 (81\%) & 0.65 (82\%) & 0.68 (86\%) & 0.71 (89\%) & 0.59 (75\%) \\
Arrhythmia & 0.75 & 0.64 (86\%) & 0.53 (71\%) & 0.56 (77\%) & 0.62 (84\%) & 0.63 (86\%) & 0.63 (86\%) & 0.64 (86\%) & 0.65 (89\%) & 0.59 (80\%) \\
Vascular & 0.83 & 0.76 (92\%) & 0.56 (68\%) & 0.62 (75\%) & 0.68 (82\%) & 0.70 (84\%) & 0.69 (83\%) & 0.70 (84\%) & 0.74 (89\%) & 0.63 (76\%) \\
CV Risk & 0.83 & 0.69 (83\%) & 0.50 (60\%) & 0.56 (67\%) & 0.65 (78\%) & 0.70 (84\%) & 0.69 (83\%) & 0.68 (82\%) & 0.67 (80\%) & 0.57 (69\%) \\
Mental & 0.81 & 0.63 (78\%) & 0.50 (62\%) & 0.55 (69\%) & 0.59 (74\%) & 0.61 (77\%) & 0.65 (81\%) & 0.63 (79\%) & 0.68 (85\%) & 0.60 (75\%) \\
Sleep & 0.76 & 0.63 (83\%) & 0.50 (66\%) & 0.52 (70\%) & 0.62 (82\%) & 0.66 (87\%) & 0.66 (88\%) & 0.65 (86\%) & 0.64 (85\%) & 0.57 (76\%) \\
Hepatorenal & 0.72 & 0.62 (87\%) & 0.53 (74\%) & 0.56 (79\%) & 0.60 (84\%) & 0.61 (86\%) & 0.62 (87\%) & 0.62 (87\%) & 0.61 (86\%) & 0.58 (81\%) \\
Musculoskeletal & 0.76 & 0.68 (89\%) & 0.50 (66\%) & 0.53 (71\%) & 0.63 (83\%) & 0.65 (85\%) & 0.64 (84\%) & 0.61 (81\%) & 0.61 (81\%) & 0.56 (74\%) \\
\midrule
Sex & 1.00 & 0.81 (81\%) & 0.50 (50\%) & 0.61 (62\%) & 0.77 (77\%) & 0.82 (83\%) & 0.82 (83\%) & 0.83 (83\%) & 0.82 (83\%) & 0.66 (67\%) \\
Age 18--39 & 0.96 & 0.86 (90\%) & 0.52 (55\%) & 0.57 (60\%) & 0.72 (75\%) & 0.75 (79\%) & 0.78 (82\%) & 0.79 (83\%) & 0.79 (83\%) & 0.65 (68\%) \\
Age 40--59 & 0.94 & 0.78 (83\%) & 0.50 (53\%) & 0.59 (63\%) & 0.67 (71\%) & 0.72 (77\%) & 0.74 (79\%) & 0.73 (78\%) & 0.74 (78\%) & 0.63 (67\%) \\
Age 60+ & 0.99 & 0.89 (90\%) & 0.50 (51\%) & 0.69 (70\%) & 0.80 (81\%) & 0.79 (80\%) & 0.81 (82\%) & 0.82 (83\%) & 0.82 (83\%) & 0.62 (63\%) \\
\midrule
BMI $<$18.5 & 0.95 & 0.85 (90\%) & 0.50 (53\%) & 0.73 (78\%) & 0.78 (83\%) & 0.78 (83\%) & 0.75 (79\%) & 0.78 (83\%) & 0.78 (83\%) & 0.66 (70\%) \\
BMI 18.5--25 & 0.91 & 0.81 (89\%) & 0.54 (60\%) & 0.56 (62\%) & 0.66 (73\%) & 0.67 (74\%) & 0.74 (81\%) & 0.70 (77\%) & 0.70 (78\%) & 0.63 (70\%) \\
BMI 25--30 & 0.78 & 0.57 (73\%) & 0.50 (64\%) & 0.52 (68\%) & 0.56 (72\%) & 0.58 (74\%) & 0.58 (75\%) & 0.59 (77\%) & 0.60 (77\%) & 0.58 (75\%) \\
BMI 30+ & 0.93 & 0.84 (90\%) & 0.53 (56\%) & 0.58 (63\%) & 0.72 (77\%) & 0.76 (82\%) & 0.77 (83\%) & 0.77 (83\%) & 0.77 (83\%) & 0.64 (69\%) \\
\midrule
A+ & 0.56 & 0.53 (94\%) & 0.50 (90\%) & 0.50 (91\%) & 0.51 (92\%) & 0.52 (93\%) & 0.52 (93\%) & 0.52 (94\%) & 0.52 (93\%) & 0.52 (92\%) \\
A$-$ & 0.59 & 0.54 (95\%) & 0.50 (86\%) & 0.53 (93\%) & 0.53 (93\%) & 0.54 (94\%) & 0.53 (93\%) & 0.53 (94\%) & 0.52 (91\%) & 0.51 (91\%) \\
B+ & 0.58 & 0.55 (95\%) & 0.51 (89\%) & 0.51 (90\%) & 0.53 (93\%) & 0.54 (93\%) & 0.54 (94\%) & 0.55 (96\%) & 0.55 (95\%) & 0.52 (91\%) \\
B$-$ & 0.51 & 0.51 (107\%) & 0.50 (100\%) & 0.48 (101\%) & 0.51 (106\%) & 0.49 (101\%) & 0.49 (103\%) & 0.48 (100\%) & 0.49 (103\%) & 0.50 (104\%) \\
AB+ & 0.54 & 0.49 (94\%) & 0.50 (98\%) & 0.51 (99\%) & 0.52 (100\%) & 0.51 (96\%) & 0.52 (100\%) & 0.52 (99\%) & 0.52 (99\%) & 0.53 (101\%) \\
O+ & 0.58 & 0.52 (92\%) & 0.51 (89\%) & 0.51 (91\%) & 0.53 (93\%) & 0.53 (94\%) & 0.53 (94\%) & 0.53 (94\%) & 0.53 (95\%) & 0.52 (92\%) \\
O$-$ & 0.54 & 0.50 (94\%) & 0.50 (93\%) & 0.52 (96\%) & 0.51 (96\%) & 0.51 (95\%) & 0.52 (97\%) & 0.51 (96\%) & 0.51 (96\%) & 0.50 (94\%) \\
\bottomrule
\end{tabular}
\end{table*}